\documentclass{article}
\usepackage{amssymb}

\usepackage[final]{corl_2025} 

\usepackage{amsmath,amssymb,amsfonts}
\usepackage{algorithmic}
\usepackage{graphicx}
\usepackage{textcomp}
\usepackage{xcolor}
\usepackage{hyperref}
\usepackage[ruled,linesnumbered]{algorithm2e}
\SetKwInput{KwData}{Input}
\SetKwInput{KwResult}{Output}

\usepackage{graphicx}
\usepackage{xspace}
\usepackage{todonotes}

\newcommand{\algoName}{\textbf{ARCH}\xspace}

\usepackage{comment}
\usepackage[numbers]{natbib}
\usepackage{booktabs}
\usepackage{array}
\usepackage{adjustbox}
\usepackage{caption}
\usepackage{multirow}  
\usepackage{colortbl}
\usepackage{float}
\usepackage{enumitem}
\usepackage{subcaption}

\title{ARCH: Hierarchical Hybrid Learning for Long-Horizon Contact-Rich Robotic Assembly}

\author{
  \textbf{Jiankai Sun}$^{1}$~~ Aidan Curtis$^{2}$~~ Yang You$^{1}$~~ Yan Xu$^{3}$~~ Michael Koehle$^{4}$~~ Qianzhong Chen$^{1}$\\\textbf{Suning Huang$^{1}$~~ Leonidas Guibas$^{1}$~~ Sachin Chitta$^{4}$~~ Mac Schwager$^{1}$~~ Hui Li$^{4}$}\\
  $^1$Stanford University~~ $^2$MIT~~ $^3$University of Michigan~~ $^4$Autodesk Research\\
  \texttt{\{jksun@,yangyou@,qchen23@,suning@,schwager@,guibas@cs.\}stanford.edu} \\ 
  \texttt{curtisa@mit.edu}~~
  \texttt{yxumich@umich.edu}~~\\
  \texttt{\{michael.koehle,sachin.chitta,hui.xylo.li\}@autodesk.com} \\ 
}

\begin{document}
\maketitle


\begin{abstract}
Generalizable long-horizon robotic assembly requires reasoning at multiple levels of abstraction. While end-to-end imitation learning (IL) is a promising approach, it typically requires large amounts of expert demonstration data and often struggles to achieve the high precision demanded by assembly tasks.
Reinforcement learning (RL) approaches, on the other hand, have shown some success in high-precision assembly, but suffer from sample inefficiency, which limits their effectiveness in long-horizon tasks.
To address these challenges, we propose a hierarchical modular approach, named \textbf{A}daptive \textbf{R}obotic \textbf{C}ompositional \textbf{H}ierarchy (\algoName), which enables long-horizon, high-precision robotic assembly in contact-rich settings. 
\algoName employs a hierarchical planning framework, including a low-level primitive library of parameterized skills and a high-level policy. 
The low-level primitive library includes essential skills for assembly tasks, such as grasping and inserting. 
These primitives consist of both RL and model-based policies.
The high-level policy, learned via IL from a handful of demonstrations, without the need for teleoperation, selects the appropriate primitive skills.
We extensively evaluate our approach in simulation and on a real robotic manipulation platform. 
We show that \algoName generalizes well to unseen objects and outperforms baseline methods in terms of success rate and data efficiency. More details
are available at: \url{https://long-horizon-assembly.github.io}.

\end{abstract}

\keywords{Robotic Assembly, Long-horizon Assembly, Hybrid Learning}

\section{Introduction}
\label{sec:intro}
\begin{figure}
    \centering
    \includegraphics[width=0.85\linewidth]{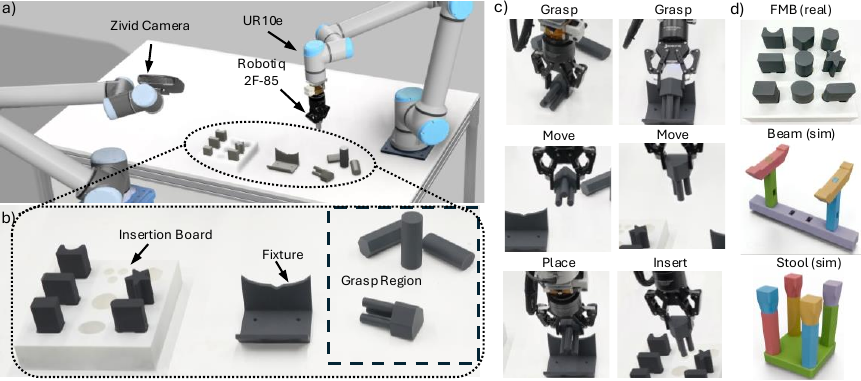}
    \captionof{figure}{
    a) Workcell setup: a camera captures RGB-D images while a robot equipped with a force-torque sensor is used for the manipulation tasks. b) Task example: objects are randomly placed within the Grasp Region on the table. The robot must insert them into the correct receptacle in the Insertion Board, adjusting their orientation with the Fixture if necessary. c) Examples of the library of RL and model-based action primitives. d) Evaluation tasks: FMB Assembly, 5-part Beam Assembly, and 9-part Stool Assembly.
    }
    \label{fig:teaser}
    \vspace{-18pt}
\end{figure}
The manufacturing industry is increasingly turning to robotic systems for assembly tasks, driven by the need for greater precision, consistency, and efficiency~\cite{cohen2019assembly, christensen2021roadmap}. Nevertheless, long-horizon, contact-rich, and high-precision robotic assembly tasks continue to pose significant challenges in robotic automation, as these tasks demand sophisticated learning and object interaction capabilities that go beyond traditional approaches~\cite{ghallab2016automated}.
Current industrial robotic applications to assembly and manufacturing are largely engineered for a specific task and struggle with adapting to varied assembly scenarios and component variations. Ideally, such systems should be capable of handling high-precision contact-rich operations~\cite{inoue2017deep} and generalizing to long-horizon new scenarios from limited demonstrations ~\cite{wang2023mimicplay,huang2025particleformer} using multimodal inputs such as visual and force-torque feedback~\cite{morgan2021vision,zhang2024dynamic}. 

Toward this goal, recent advances in learning-based solutions have started to tackle some of these problems.
End-to-end imitation learning (IL) from human demonstrations has made encouraging progress~\cite{chi2023diffusion,zang2023peg,wang2021robotic}, but usually requires a significant amount of expert demonstration data~\cite{heo2023furniturebench}, which can be costly to collect, and often struggles with high-precision tasks. While reinforcement learning (RL) can drive specific assembly operations, such as high-precision insertion, it usually falls short on more complicated long-horizon tasks~\cite{schoettler2020deep,luo2019reinforcement,sun2021adversarial,huang2024dittogym,chen2025gradnav}, where training faces challenges of sparse rewards and an exploding sampling space due to the larger long-horizon footprint. Alternatively, naively sequencing atomic actions or behaviors can lead to compounding errors, diminishing overall system performance~\cite{sun2022plate}.

To address these challenges, we propose a hierarchical approach for robotic assembly, especially targeted at long-horizon, contact-rich, high-precision settings. Our method adopts a hierarchical framework that consists of a library of parameterized low-level primitive skills (e.g., grasp, insert) and a high-level policy that sequences these primitives based on task context.
To build robust and adaptable low-level primitives, we leverage both classical motion planning (MP) algorithms and RL policies. This hybrid MP-RL strategy enables the system to leverage prior knowledge for efficient execution of routine motions while remaining flexible enough to handle contact-rich and uncertain dynamics. The RL components are trained in simulation using domain randomization, enabling feasible transfer to real-world execution.
At the high level, the policy operates over a compact action space, enabling efficient learning from a handful of human demonstrations via imitation learning, thereby alleviating the data collection burden common in end-to-end imitation learning pipelines.

In summary, our main contributions are as follows:
\begin{itemize}[leftmargin=0em]
    \item We introduce \textbf{A}daptive
\textbf{R}obotic \textbf{C}ompositional \textbf{H}ierarchy (\algoName), a hierarchical framework designed to tackle the challenging problem of \emph{long-horizon robotic assembly}. 
    \item 
    We develop a hybrid low-level skill library that includes both MP algorithms and RL policies for efficient, adaptive, and high-precision behaviors in contact-rich assembly scenarios. Additionally, we design a high-level policy based on a Diffusion Transformer (DiT), trained via imitation learning with only a handful of demonstrations, to efficiently sequence and deploy these primitive skills.
    \item  We evaluate our framework on three long-horizon robotic assembly tasks, each involving the assembly of 4 to 9 objects. Experiments in both simulation and the real world show that, despite being trained on a single object, our approach generalizes to unseen objects and outperforms baseline methods in terms of both success rate and data efficiency.
\end{itemize}
 
\section{Related Work}
\label{sec:related_work}

\subsection{Task and Motion Planning}
Task and Motion Planning (TAMP)~\cite{kaelbling2011now, kaelbling2017backchain, garrett2021integrated} is an effective approach for long-horizon manipulation problems as it can resolve temporally dependent constraints through hybrid symbolic-continuous reasoning~\cite{sun2021neuro,huang2021learning}. These plans may involve regrasping~\cite{adubredu2022optimalconstrainedtaskplanning}, clearing obstructing objects~\cite{m0m}, or moving to gather information~\cite{curtis2024partiallyobservabletaskmotion}. Despite these abilities, such methods typically require hand-designed symbolic transition functions and continuous parameter samplers. Additionally, TAMP methods are computationally expensive, which limits their ability to handle failures in dynamic tasks. 
We argue that the symbolic transition functions and samplers can be replaced by a learnable high-level policy to address the issues mentioned above. 

\subsection{Learning for Long-Horizon Manipulation}
There is a large body of research in learning for long-horizon manipulation tasks. Diffusion policies~\cite{chi2023diffusion, chi2024umi, ze20243d} have been shown to be a powerful tool for addressing complex manipulation tasks by leveraging their generative and multi-modal capabilities. However, they require a large amount of training data, i.e., human demonstrations, to learn effective end-to-end policies for complex, long-horizon tasks. Therefore, a significant amount of research has adopted the divide-and-conquer mindset and tackled the task by decomposing it into easier and reusable subtasks, instead of learning an entire task with a single policy.

Skill decomposition and chaining~\cite{huang2024mentor,chen2025grad,konidaris2009skill, konidaris2012lfd, agia2022taps, lee2019composing, lee2021adversarial, clegg2018dress} is a promising way to synthesize long-horizon and complex behaviors by sequentially chaining previously learned simpler skills through transition functions. Mishra et al.~\cite{mishra2023generative} trains individual skill diffusion models as action primitives and combine them at test time, when learned distributions of the skills are linearly chained to solve for a long-horizon goal during evaluation. Mao et al.~\cite{mao2024dexskills} learns primitive skills from human demonstrations based on haptic feedback and segments a long-horizon task into a sequence of skills for dexterous manipulation. Chen et al.~\cite{chen2023seqdex} introduces a bi-directional optimization framework that chains RL-trained sub-policies together using a transition feasibility function for long-horizon dexterous manipulation. 

\subsection{Hierarchical Modeling in Robotics}
Hierarchical approaches typically offer policies at varying abstraction levels~\cite{gao2024prime,dalal2021accelerating,yang2024learning,sun2021hiabp,ye2024reasoning,zhang2025synapseroute}. Xian et al. \cite{xian2023chained} and Ma et al. \cite{ma2024hierarchicaldiffusion} propose to learn a high-level policy from demonstrations to predict end-effector key poses and a low-level diffusion-based trajectory generator for connecting these key poses to achieve the final goal. MimicPlay~\cite{wang2023mimicplay} learns a high-level plan from human videos of manipulating objects and low-level visuomotor controls from teleoperated demonstrations on the real robot. MAPLE~\cite{nasiriany2022maple} enhances standard RL algorithms by incorporating a predefined library of behavioral primitives. The most relevant work to our approach uses hierarchical policy decomposition for a multi-stage cable routing task \cite{luo2024multi}. They define both scripted and imitation-learned low-level primitives, and train a high-level policy to select among them. In contrast, we incorporate both model-based and model-free low-level primitives, and learn a high-level policy that selects and instantiates parameterized primitives. Our RL policy for the contact-rich insertion primitive is trained in simulation and transferred to the real world via zero-shot sim-to-real deployment.

\section{Method}
\label{sec:method}
\begin{figure}
    \centering
    \includegraphics[width=\linewidth]{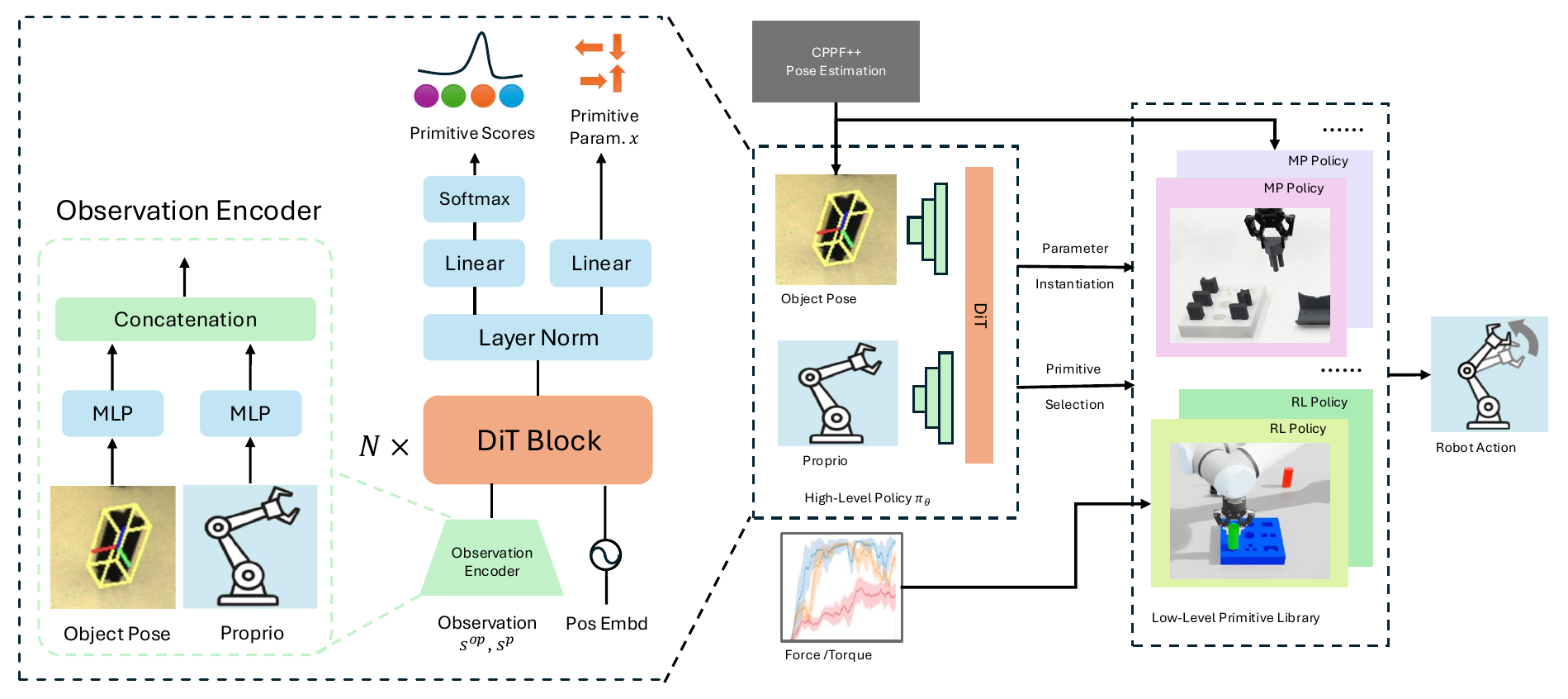}
    \caption{We propose a hierarchical framework for long-horizon robotic assembly. The high-level policy $\pi_{\theta}$, trained via imitation learning, takes as input the object pose from a pose estimator and the robot’s proprioceptive state, and outputs a categorical distribution over low-level primitives and their parameters. The selected primitive is then executed by a low-level policy, which is either reinforcement learning (RL)-based or motion planning (MP)-based. Specifically, we train RL policies in simulation for contact-rich skills, such as `insert', while MP policies are used for primitives operating in free space, such as `move'.
}
    \label{fig:framework}
\end{figure}
In this work, we model the long-horizon robot assembly task as a parameterized-action Markov decision process (PAMDP)~\cite{pamdp}. A PAMDP problem consists of the tuple $\langle \mathcal{S}, \mathcal{A}, P, R, \gamma \rangle$ where $\mathcal{S}$ is the continuous state space, $\mathcal{A}$ is a set of discrete action primitives $\{a_1, a_2, ..., a_k\}$, each with $m_a$ parameters $X_a \subseteq{\mathbb{R}^{m_a}}$, $P(s, a, s')$ is the probability of transitioning to state $s'$ when taking action primitive $a$ in state $s$, $R(s, a)$ is the reward from taking action primitive $a$ in state $s$, and $\gamma$ is the discount factor. Each parameterized action primitive $a$ can be instantiated as either a model-based policy or an RL policy.
Our goal in a PAMDP is to find a policy $\pi(a, x | s)$ that minimizes the discrepancy between the learned policy and the expert demonstrations that maximize reward under the PAMDP.

Solving a PAMDP involves not only selecting the best discrete primitive but also optimizing over the parameter space for each primitive at each state. In this work, we fix a set of primitives that are building blocks of manipulation for assembly tasks (See Section~\ref{sec:primitives}) and attempt to learn the high-level policy $\pi_{\theta}(a, x|s)$ as a neural network with parameters $\theta$. This results in a hierarchical framework, shown in Figure~\ref{fig:framework}. In this section, we describe our assumptions, the low-level primitive library, the high-level policy architecture, the data collection pipeline, and the pose estimation module.

We make the following assumptions in this work: (1) the CAD models of the assembly parts are known, as is common in industrial settings; (2) the base component of each assembly, as well as the fixture used for object reorientation, is fixed to the table. Under these assumptions, a single pose estimation step—combined with the known CAD models—is sufficient to determine the target insertion poses for each object with minimal error. Note that assumption (2) can be relaxed by performing continuous pose estimation, i.e., repeatedly estimating object poses throughout the task to account for potential shifts or disturbances.

\subsection{Low-level Primitive Library: Modeling Basic Skills with Motion Planning and Reinforcement Learning}\label{sec:primitives}
We focus on providing agents with a library of flexible primitives that act as foundational components for high-precision, contact-rich robotic assembly tasks. The hierarchical decision-making system operates independently of the specific implementations of these primitives, which may include either closed-loop, learning-based skills or model-based motion planners. 
Regardless of how they function internally, it is essential that the primitives are adaptable to varying behaviors, which is why we introduce parameters to customize each primitive. 
These parameters typically have clear semantics, such as an object index or a 6-DoF end-effector pose for a grasping primitive. While these primitives offer flexibility, we acknowledge that they are not exhaustive and can be extended. 

\noindent\texttt{GRASP$(o, p_g)$}: The robot moves its end-effector to pre-grasp pose $p_g\in\mathbf{SE}(3)$ and closes its gripper to grasp object $o$. A motion planner is used for its execution.

\noindent\texttt{PLACE$(o, p_p)$}: The robot has object $o$ in grasp and moves its end-effector to pre-place pose $p_p\in\mathbf{SE}(3)$, and opens its gripper to place object $o$. A motion planner is used for its execution.

\noindent\texttt{MOVE$(g_m)$}: The robot moves its end-effector to pose $g_m\in\mathbf{SE}(3)$. A motion planner is used for its execution.

\noindent\texttt{INSERT$(o, g)$}: The robot has object $o$ in grasp and inserts it into its goal pose $g\in\mathbf{SE}(3)$. An RL policy $\pi_L(\phi)$ is employed for two primary reasons: first, insertion goal poses are subject to inaccuracies from pose estimation errors; second, insertion tasks inherently involve complex, contact-rich interactions between the object and its receptacle. We train an RL policy in simulation~\cite{mittal2023orbit} using PPO~\cite{schulman2017proximal}. Its observation space includes end-effector (EE), force-torque (FT), and EE pose relative to the goal pose; its action is EE velocity $u_t$; and its reward $\tilde{r}(s^p_t, u_t, g)$ is the negative distance between the current pose $s^p_t$ and the goal pose $g$.

The objective of the goal-conditioned \texttt{INSERT} primitive is to reach the goal pose $g$ that maximizes the expectation of the cumulative return, as Equation~\ref{eq:insert_loss} shows:
\begin{equation}
\label{eq:insert_loss}
   \mathcal{J}(\phi) = \mathbb{E}\left[\sum_t\gamma^t\tilde{r}(s^p_t, u_t, g)\right].
\end{equation}

\subsection{High-level Policy: Composing Primitives via Imitation} 
As shown in Figure~\ref{fig:framework}, the high-level policy $\pi_\theta (a, x | s)$ takes pose information obtained from pose estimation (Section~\ref{sec:pose_est}) for a given object $o$ and robot proprioception as inputs $s$, to select the appropriate primitive $a$ from the low-level primitive library and instantiates it with parameters $x$ for low-level control. Parameters $x$ contains the object index $o$ and pose information obtained from pose estimation. We collect human demonstration data (Section~\ref{sec:data}) and train the high-level policy via imitation learning. 

We modify the Diffusion Transformer (DiT)~\cite{peebles2023scalable} architecture as the backbone of our high-level policy, considering that its strong sequential data handling capability is suitable for handling a history of previous states and actions. The DiT outputs softmax scores for each primitive and the continuous action parameter for the primitive with the highest score. These two functions are supervised using cross-entropy loss and MSE loss, respectively. We employ DiT blocks with adaptive LayerNorm-Zero conditioning. We use the Robomimic observation encoder~\cite{robomimic2021} to extract features from object pose ($s^{op}$) and proprioceptive pose ($s^p$). The goal of imitation learning is to find a parameterized policy $\pi_\theta$ that can maximize the likelihood function based on the currently collected demonstration data $\mathcal{D}=\{(\mathbf{s, a, x)}\}$:
\begin{equation}
    \label{eq:hl_loss}
    \theta = \arg\max_\theta \mathbb{E}_{\mathbf{s, a, x}\sim \mathcal{D}}\pi_\theta(\mathbf{a, x|s}).
\end{equation}

\subsection{Data Collection}\label{sec:data}

As our low-level primitives are implemented with either RL policies or MP policies, no demonstration data is needed. In this section, we focus on the data collection process for the high-level IL policy. Our hierarchical framework abstracts away the motion details from the demonstrator and allows them to demonstrate by selecting primitives and parameters instead of teleoperation. More specifically, from the library of pre-defined action primitives (Section~\ref{sec:primitives}), a human demonstrator uses a keyboard to select the most appropriate primitive and its parameters to complete the assembly task at hand. 
For example, to initiate the \texttt{GRASP} primitive of object $o$, the demonstrator selects the primitive index and the object index. The pose estimation module (Section~\ref{sec:pose_est}) estimates the object pose $p_o$ and the continuous parameter, pre-grasp pose $p_g$, is calculated by adding a fixed offset from $p_o$. $p_g$ is then passed to the motion planner to execute the primitive. To initiate the \texttt{INSERT} primitive of object $o$, the demonstrator selects the primitive index and the object index. As described in the assumptions, the insertion goal pose $g$ for each object is known with a small margin of error. A pre-trained RL policy is used with $g$ as its goal pose.

The dataset denoted as $\mathcal{D} = \{(\mathbf{s},\mathbf{a},\mathbf{x})\}$, consists of sensor observations $\mathbf{s}$, primitive indexes $\mathbf{a}$, and corresponding primitive parameters $\mathbf{x}$. The dataset consists of two types of demonstrations: “successful” and “recovery” trials. In half of the trials, the demonstrator successfully inserts the object, while in the other half, failures occur, followed by recovery actions demonstrated by the operator. These recovery trials are crucial to making the learned policy more robust to errors. 

In summary, we allow the demonstrator to select the discrete primitive and parameter to execute while the continuous input parameter comes from pose estimation. This is a novel way to collect demonstration data as it is often difficult and time-consuming to collect teleoperated demonstrations for high-precision tasks. 

\subsection{Pose Estimation}\label{sec:pose_est}
As shown in Figure~\ref{fig:framework}, object pose is needed by the high-level policy and also serves as the input parameter to the low-level primitive skills, such as \texttt{grasp}. We integrate the pose estimation method CPPF++~\cite{you2022cppf,you2024cppf++}, which demonstrates strong generalization from simulation to real-world scenarios. Specifically, given a 2D RGB-D image $\mathcal{I}$, the model produces an accurate 6D pose $\xi \in \mathfrak{se}(3)$, where the first three components represent rotation and the last three correspond to translation. While the original CPPF++ is designed for category-level pose estimation, we adapt it for instance-level pose estimation by using distinct CAD models as different ``categories''. To enhance precision, inspired by the Iterative Closest Point (ICP)~\cite{arun1987least} algorithm, we propose a post-optimization procedure conducted in the tangent space of the Lie group. 

Specifically, we compute the one-way Chamfer distance from the object-masked point cloud to the full CAD model, transformed by the current pose $\xi$: 
\begin{equation}
    L_{CD}(\xi) = \sum_i\min_j \|T_\xi(p_i) - p^*_j\|_2,
\end{equation}
where $p^*_j$ is the $j$-th point on the back-projected point cloud obtained from the predicted masks, $p_i$ is the $i$-th point on the object mesh, and $T_\xi$ is the rigid transformation that aligns the canonical object point with the scene, defined as $T_\xi(p_i) = R\cdot p_i + t$. We utilize the one-way Chamfer distance due to self-occlusion in the observation and optimize $\xi$ iteratively using the Lietorch~\cite{teed2021tangent} library to obtain a refined pose.
Figure~\ref{fig:pose_estimation} in the Appendix presents some qualitative pose detection results. The high accuracy of the pose estimator enables us to have a high success rate in terms of both the low-level MP policies and the high-level policy.

\section{Experiments}
\label{sec:exp}

\subsection{Task Description and Setup}\label{sec:task}
Recent advancements in robotic learning have also highlighted the need for benchmarks that effectively balance generalization and the complexity of manipulation tasks. The Functional Manipulation Benchmark (FMB)~\cite{luo2024fmb} aims to bridge this gap by defining functional manipulation as a series of relevant behaviors, such as grasping, repositioning, and physically interacting with objects. Building on FMB, we want to extend its emphasis on contact-rich dynamics and object diversity.
For these reasons, we use the multi-peg assembly task introduced in FMB, along with two more challenging custom-designed multi-part assembly tasks, to showcase the generalizability of our approach.

We focus on three assembly tasks: FMB Multi-peg Assembly in~\cite{luo2024fmb}, 5-part Beam Assembly, and 9-part Stool Assembly, as shown in Figure~\ref{fig:teaser}. 
FMB Assembly is tested on the real robot, whereas the other two assemblies are in simulation only. 
For FMB Assembly, 9 objects of different shapes must be inserted into the board. The robot must grasp the object that is randomly placed on the table, may need to reorient and regrasp it using the fixture depending on its initial pose, and then insert it into the board. 
More task and setup details can be found in the Appendix.
We evaluate single-step and long-horizon categories and examine generalization to unseen objects and tasks. 

\subsection{Baselines}
We evaluate two end-to-end (E2E) baselines: 
1) \textbf{E2E IL}: Diffusion Policy (DP)~\cite{chi2023diffusion}, a goal-conditioned imitation learning framework that leverages a diffusion model for generating diverse action trajectories. Teleoperated demonstration data is collected in the real world.
2) \textbf{E2E RL}: PPO~\cite{schulman2017proximal},  a widely-used RL method that optimizes policy performance through proximal updates, balancing exploration and exploitation while ensuring stable training. Training is conducted in the IsaacLab simulation environment~\cite{mittal2023orbit}.

We also evaluate two hierarchical baselines:
1) \textbf{MimicPlay}~\cite{wang2023mimicplay} learns high-level latent plans from human play data to guide low-level visuomotor control. Teleoperated demonstration data is collected in the real world.
2) \textbf{Luo et al.}~\cite{luo2024multi} propose a Multi-Stage Cable Routing approach through hierarchical imitation learning. Teleoperated demonstration data for both low-level and high-level policies is collected in the real world. We adapt the above methods to our task. 

\begin{table}[]
    \centering
    \caption{Success rate SR(\%) of the multi-stage FMB Assembly of Hexagon, Beam Assembly, and Stool Assembly, comparing our method with baseline methods after training with $10$ demonstrations, with the human oracle being the upper bound.
    }
    \begin{tabular}{l|cc|cc|cc}
    \toprule
Methods & \multicolumn{2}{c|}{FMB (real)} & \multicolumn{2}{c|}{Beam (sim)} & \multicolumn{2}{c}{Stool (sim)} \\
Metrics       & SR (\%) $\uparrow$ & SPL $\uparrow$ & SR (\%) $\uparrow$ & SPL $\uparrow$ & SR (\%) $\uparrow$ & SPL $\uparrow$ \\
    \midrule
    E2E RL~\cite{schulman2017proximal}     & 0 & 0.00 & 0 & 0.00 & 0 & 0.00 \\
    E2E IL~\cite{chi2023diffusion} & 0 & 0.00 & 0 & 0.00 & 0 & 0.00 \\
    MimicPlay~\cite{wang2023mimicplay} & 20 & 0.16 & 15 & 0.12 & 10 & 0.08 \\
    Luo et al.~\cite{luo2024multi} & 25 & 0.19& 15 & 0.14 & 20 & 0.18 \\
    \algoName (Ours) & 55 & 0.51 & 55 & 0.51 & 45 & 0.44 \\
    \hline
    Human Oracle~\cite{luo2024fmb} & 65 & 0.55 & 60 & 0.62 & 50 & 0.49 \\
    \bottomrule
    \end{tabular}
    \label{tab:quat}
\end{table}

\subsection{Evaluation Metrics}
A total of 20 trials are conducted for each task. For each trial, the selected object is initialized randomly in the grasp region.
For the \texttt{GRASP} primitive to succeed, the object must be securely gripped by the gripper without falling after being lifted. For the \texttt{INSERT} primitive to succeed, the object must be fully inserted into the corresponding receptacle. 

The main metric is the overall task success rate (\%). We found that failure primarily stems from grasping and insertion. Hence, we also introduce two metrics: "\% Grasped" and "\% Inserted", to measure the success rates of these two primitives.

\textbf{Success Rate (SR\%)}: percentage of successful completion of the long-horizon tasks.

\textbf{Success weighted by Path Length (SPL)}:
\begin{equation}
   \label{eq:spl}
   \textrm{SPL} = \frac{1}{N}\sum_{i=1}^NS_i\cdot \frac{l_i}{\max(p_i, l_i)},
\end{equation}
where \( N \) is the total number of trials, \( S_i \) is a binary indicator of success in trial $i$, \( l_i \) is the shortest path (number of primitives) to the goal, and \( p_i \) is the actual path length taken by the agent.

\textbf{\% Grasped}: percentage of successful grasps as a single-stage task.

\textbf{\% Inserted}: percentage of successful insertions as a single-stage task.

\begin{table}[]
    \centering
    \vspace{-10pt}
    \caption{Success rate SR(\%) of the multi-stage FMB Assembly task after training with $10$ demonstrations and success rate of the single stages by object. Our system demonstrates the ability to generalize to unseen objects.}
    \begin{tabular}{l|c|c|c|c}
    \toprule
    & Object  & SR (\%)  & \% Grasped & \% Inserted  \\
    \midrule
    \rowcolor{gray!50}
    \multirow{1}{*}{Seen} & Hexagon     & 50 & 80 & 75  \\
    \midrule
    \multirow{8}{*}{Unseen} & Star & 50 & 85 & 60 \\
    & SquareCircle & 35 & 75 & 50  \\
    & 3Prong & 80 & 90 &  90  \\
    & Circle & 75 & 95 &  80 \\
    & Oval & 55 & 85 &  70 \\
    & Arch & 40 & 65 & 65 \\
    & DoubleSquare & 40 & 70 & 60 \\
    & Rectangle & 65 & 90 & 75  \\
    \bottomrule
    \end{tabular}
    \label{tab:quat_single}
\end{table}
\subsection{Experiment Analysis}
\textbf{Comparative Performance Analysis.} First, we compare with end-to-end baselines and hierarchical approaches. For methods that utilize a low-level primitive library, we ensure they remain identical across all methods to ensure a fair comparison.
As shown in Table~\ref{tab:quat}, long-horizon tasks often involve sparse rewards, which complicates RL training. Consequently, \textbf{E2E RL}~\cite{schulman2017proximal} fails at such long-horizon assembly tasks. With diffusion models, \textbf{E2E IL}~\cite{chi2023diffusion} falls short in handling high-precision tasks and requires a large number of expert demonstrations. The hierarchical baselines, \textbf{MimicPlay}~\cite{wang2023mimicplay} and \textbf{Luo et al.}~\cite{luo2024multi}, achieve better success rate than the E2E methods, but struggle with contact-rich steps such as insertion due to the small number of demonstrations available. Finally, \textbf{Human Oracle} is not a baseline but serves as the upper bound for high-level policy. The human oracle selects primitives with near-flawless accuracy and can self-correct mistakes. Failures with this method are due to pose estimation and grasp errors.
\algoName demonstrates significantly higher success rates than baseline methods due to our hierarchical framework and the hybrid MP-RL design of the low-level library. The advantage of \algoName becomes more pronounced on the more challenging Beam Assembly and Stool Assembly tasks compared to baseline methods. By combining model-based and learning-based components, our approach achieves both high precision and flexibility. Failures with our approach are mainly due to inaccuracies in pose estimation and limitations in the RL policies. Since pose estimation and insertion policies are extensively studied areas, numerous established methods~\cite{foundationposewen2024,zhang2024dynamic} exist to enhance performance; however, exploring these improvements is beyond the scope of this work. 

\textbf{Generalization to Unseen Objects.} \hspace{0.5em} Our system has been trained exclusively with the Hexagon object, both for the RL primitive and for the high-level policy. Table~\ref{tab:quat_single} shows that our system generalizes well to unseen objects without any fine-tuning. Objects with a narrow edge, such as the SquareCircle and DoubleSquare, present more challenge during the grasping stage. Additionally, certain object shapes, such as the Star, SquareCircle, and Arch, present more challenge to pose estimation.

\textbf{Data Efficiency.} \hspace{0.5em} As Table~\ref{tab:quat} shows, \algoName achieves high success rates with merely $10$ demonstrations, whereas baseline methods would require significantly more data. This capability not only reduces the overall data collection burden but also enhances the practical applicability of our method in real-world scenarios, where data acquisition can be a limiting factor. 

\textbf{Robustness.} \hspace{0.5em} Our high-level policy exhibits robustness to single-stage failures and human disturbances via retries. For example, we observed that when the grasp primitive fails, the policy automatically triggers another grasp attempt using the new pose estimate. This failure recovery capability contributes to the higher success rate of \algoName.

\section{Conclusion}
In this paper, we introduce a hierarchical hybrid learning system for long-horizon contact-rich robotic assembly. It includes a low-level parameterized skill library and an imitation-learned high-level policy for selecting and composing primitive skills. The hierarchical structure and the pre-trained primitive skills enable our system to be data efficient for long-horizon tasks, while satisfying the high-precision requirement of assembly.

\newpage
\section{Limitations} 
\label{sec:limitations}

\textbf{Cross-Task Generalization.} Although the learned high-level policy transfers across novel objects within a given task, it remains task-specific: separate training is still required when moving from, for example, FMB Assembly to Stool Assembly. However,  training a high-level policy for a new task requires only a handful of demonstrations and no teleoperation, keeping the adaptation effort minimal.

\textbf{Object Identification Pipeline.} All objects are manually labeled and indexed prior to execution. Replacing this offline step with a perception module—e.g., a vision-language model–based detector that auto-assigns semantic IDs—would streamline this process and improve overall efficiency.

\textbf{Scalability and Generality of the Primitive Library.}
Our approach assumes that tasks can be decomposed into a finite set of parameterized primitives. This assumption may pose scalability challenges for tasks that require fundamentally new skills. However, in practice, we observe that a relatively small library of primitives—automatically discovered from demonstrations—generalizes effectively across a wide range of long-horizon tasks. Moreover, in industrial settings, tasks generally require only a small set of primitives. In addition, our framework is modular and can readily integrate newly introduced primitives when needed, making it naturally extensible to novel tasks.

\appendix


\bibliography{references}  
\newpage
\section{Appendix}
\label{sec:appendix}

\subsection{Task Descriptions}
In addition to FMB Assembly, we introduce two more challenging long-horizon assembly tasks, both of which involve multi-stage dependencies. These dependencies make the tasks more difficult—for instance, the feet can only be inserted after the legs are correctly assembled. If the legs are not properly inserted, subsequent steps will fail.

\textbf{FMB Assembly}: This task involves assembling 9 distinct objects with varying shapes. For each object, the robot must first grasp it, then perform a sequence of reorientation actions using an environment fixture, followed by an insertion into the board. The time horizon for each object ranges from 20 to 40 seconds.

\textbf{5-part Beam Assembly}: The base part is fixed to the table. The task involves inserting two legs into the base, followed by inserting two feet onto the legs. The time horizon for each object ranges from 20 to 40 seconds. 

\begin{figure}[H]
  \centering
  \begin{tabular}{>{\centering\arraybackslash}m{0.15\textwidth}
                  >{\centering\arraybackslash}m{0.38\textwidth}
                  >{\centering\arraybackslash}m{0.38\textwidth}}
  \toprule
    \textbf{Task Name} &
    \textbf{5-part Beam Assembly} &
    \textbf{9-part Stool Assembly} \\
    \midrule
    \textbf{Initial State} &
    \begin{subfigure}[t]{\linewidth}
      \centering
      \includegraphics[trim=150 0 0 0, clip, width=0.9\linewidth]{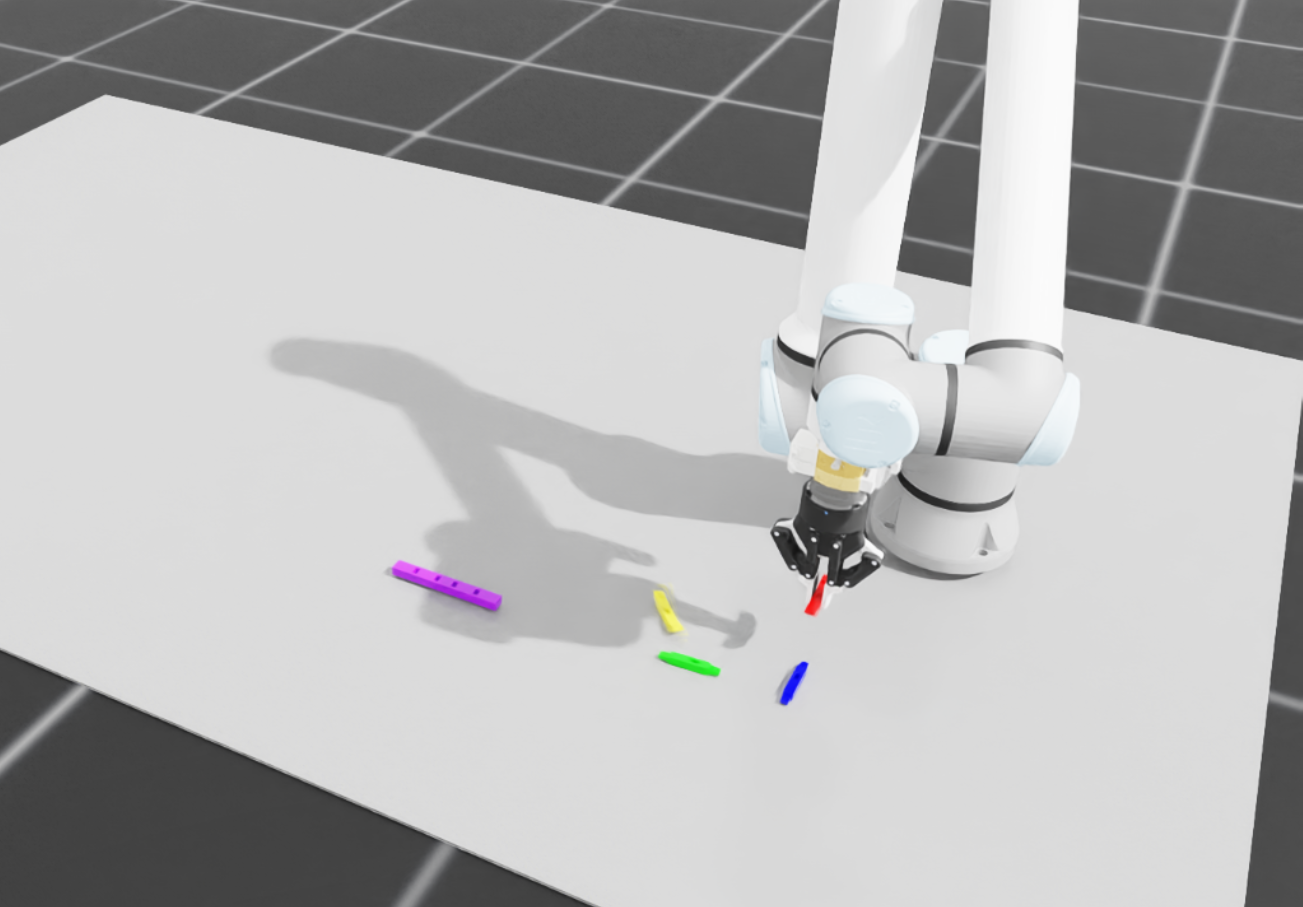}
    \end{subfigure} &
    \begin{subfigure}[t]{\linewidth}
      \centering
      \includegraphics[trim=150 0 35 0, clip, width=0.9\linewidth]{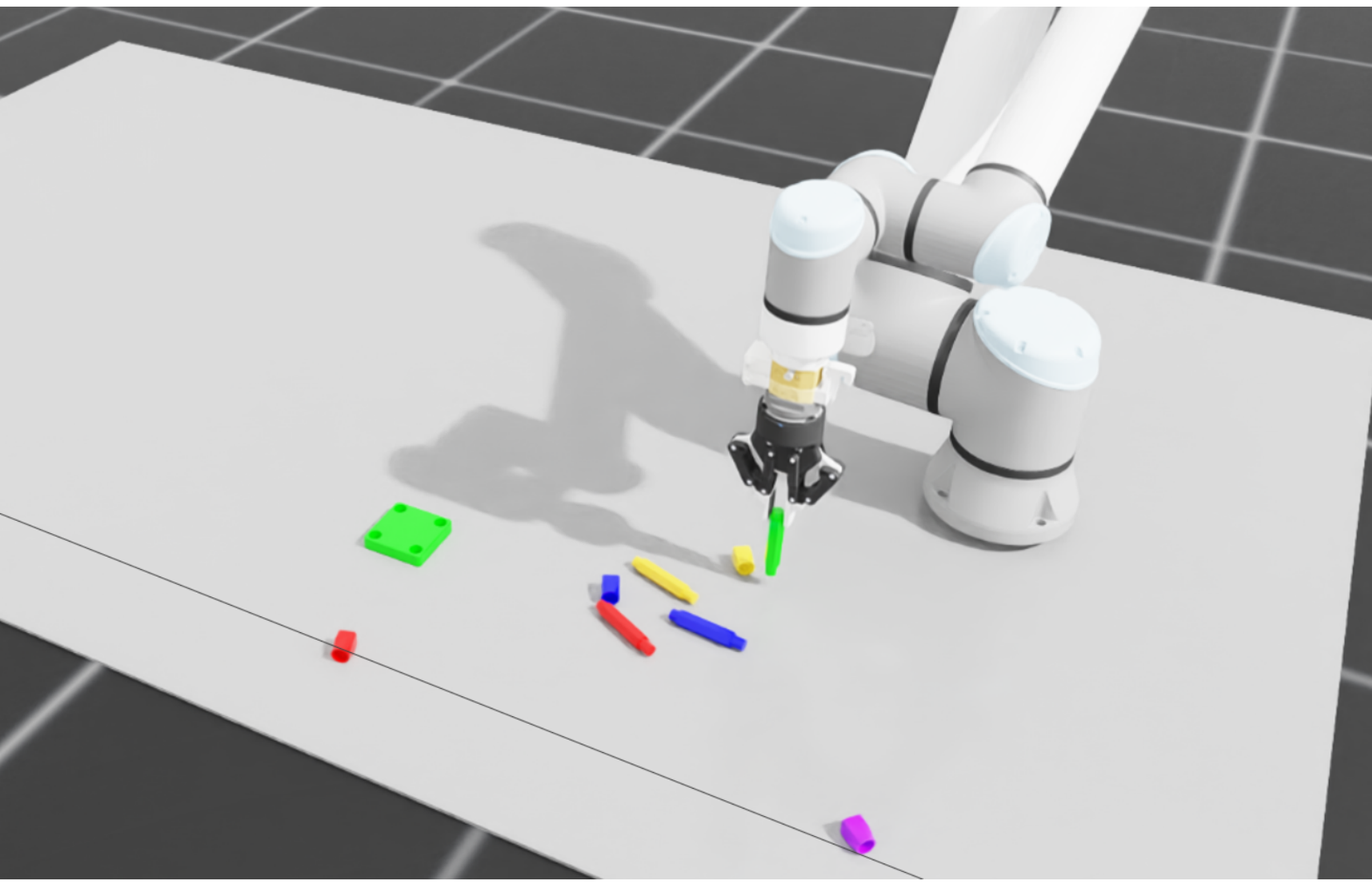}
    \end{subfigure} \\
    \textbf{Final State} &
    \begin{subfigure}[t]{\linewidth}
      \centering
      \includegraphics[trim=150 0 0 0, clip, width=0.9\linewidth]{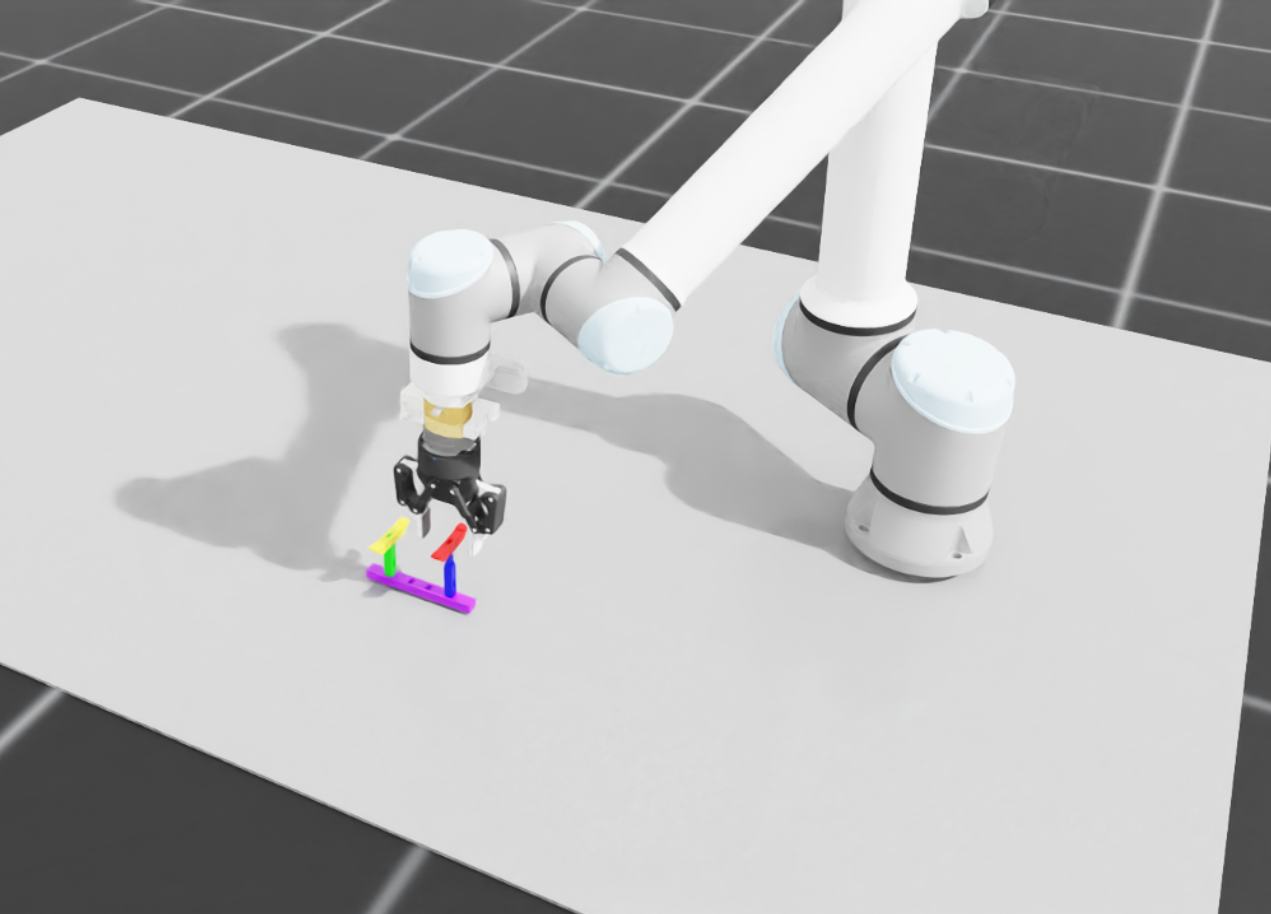}
    \end{subfigure} &
    \begin{subfigure}[t]{\linewidth}
      \centering
      \includegraphics[trim=150 0 35 0, clip, width=0.9\linewidth]{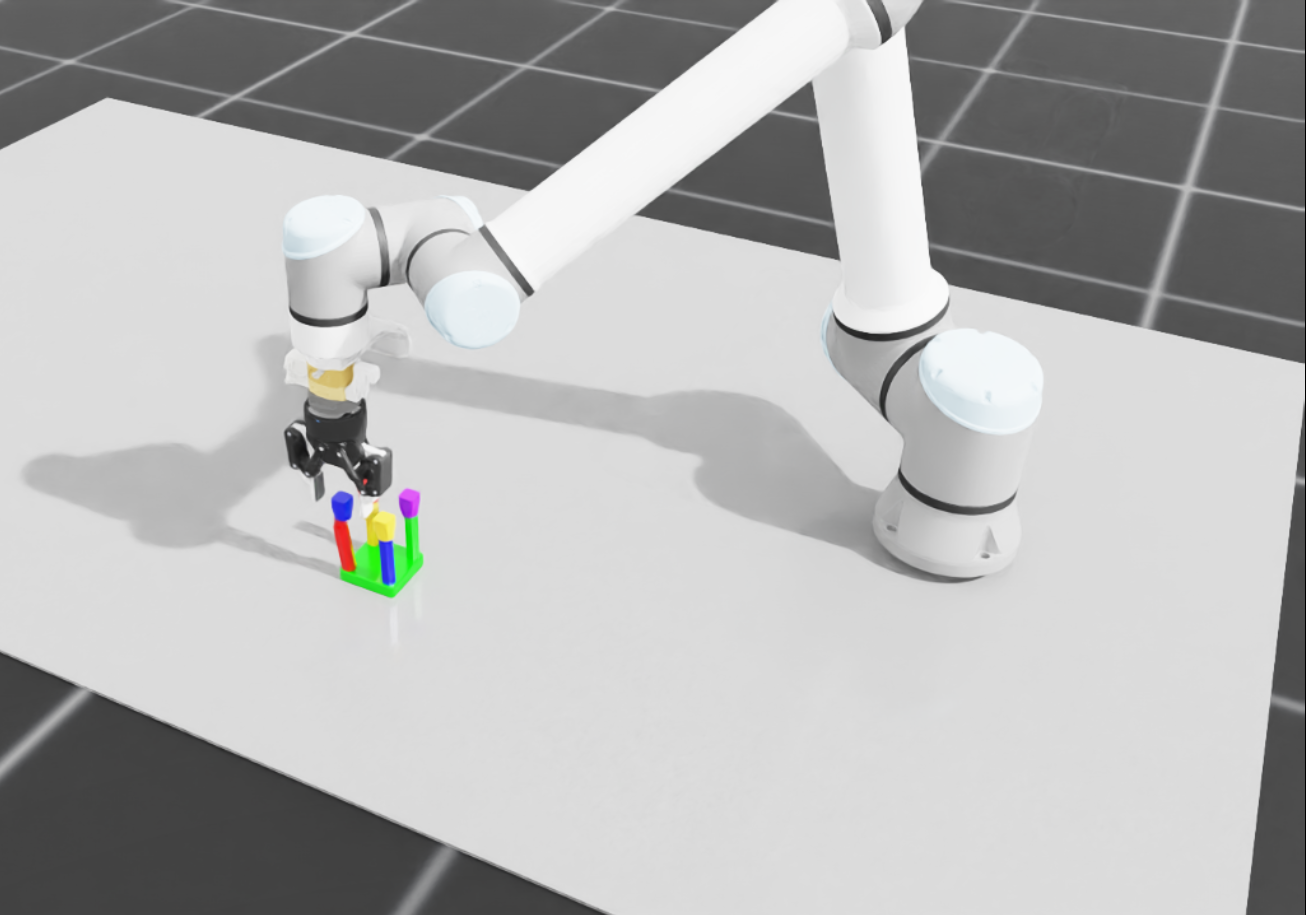}
    \end{subfigure} \\
    \bottomrule
  \end{tabular}
  \caption{Simulation setup for Beam Assembly and Stool Assembly in IsaacLab.}
  \label{fig:setup-stool}
\end{figure}

\textbf{9-part Stool Assembly}: As shown in Figure~\ref{fig:setup-stool}, the base is also fixed to the table. This task involves inserting four legs into the base, and then inserting four feet onto the legs. The increased number of objects and stages makes this task significantly more challenging and long-horizon. The time horizon for each object ranges from 20 to 40 seconds. 

For visualizations of the initial and final states of two new task, Beam Assembly and Stool Assembly, please refer to Figure~\ref{fig:setup-stool}.

\subsection{Experiment Details}
\label{sec:appendix-setup}
\noindent\textbf{Real Workcell Setup.} 
The real workcell setup includes a UR10e robotic arm and a third-person view Zivid camera (Figure~\ref{fig:teaser}). The simulation platform for RL training is built on the IsaacLab engine~\cite{mittal2023orbit}, while motion-level manipulation plans are computed using lazyPRM~\cite{lazyPRM2000bohlin}. Model training is performed on an Ubuntu machine equipped with an Intel i7 CPU and a GeForce RTX 4090 GPU. The Beam Assembly and the Stool Assembly are evaluated in simulation. 

\noindent\textbf{Hyperparameters and Computation.} 
For pose estimation, we use $512\times 512$ RGB-D image as input.
For high-level policy $\pi_\theta$, 
We adopt the Diffusion Transformer model~\cite{peebles2023scalable} for predicting action feasibility score.
Our RL primitive is trained using 1,000 parallel environments in IsaacLab. 
The maximum task horizon is 25,000 action steps, equivalent to 200 seconds of robot execution at a 125Hz control frequency.

For our DiT-based high-level policy, the number of reverse diffusion timesteps is a critical parameter that significantly impacts both the sampling time and the quality of the generated samples. We choose 128 diffusion steps, which provide the best balance for most tasks. Additional hyperparameters for the DiT network are listed in Table~\ref{tab:hyper_dit}.

\subsection{Pose Estimation Results}
\begin{figure}[H]
    \centering
    \includegraphics[width=\linewidth]{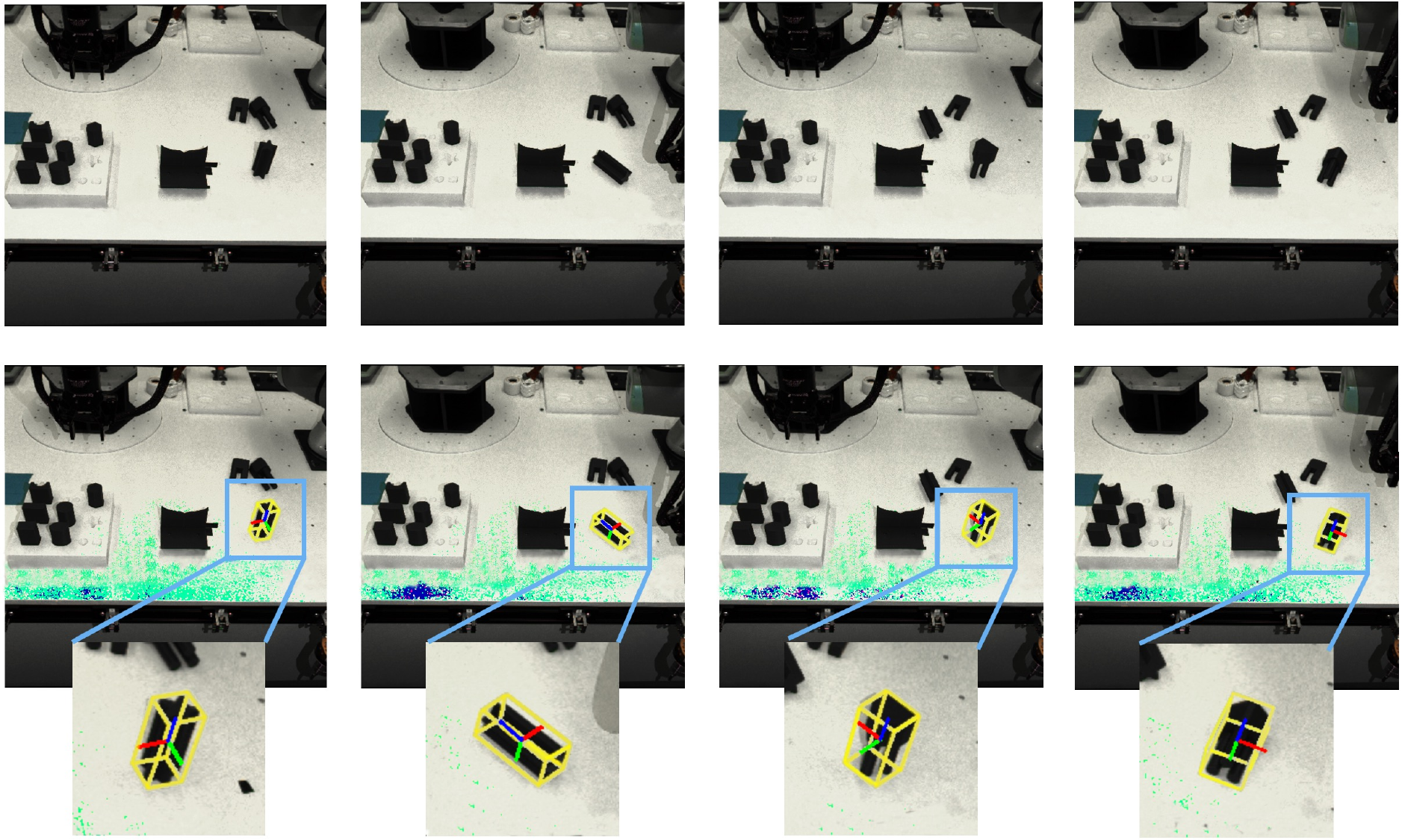}
    \caption{\textbf{Pose Estimation Examples}. Our method gives accurate pose estimation of differently shaped parts. The top row shows the input images, and the bottom row shows the pose estimation. \textcolor{red}{Red}, \textcolor{green}{green}, \textcolor{blue}{blue} colors indicate the $xyz$ axes in the canonical space, respectively.}
    \label{fig:pose_estimation}
\end{figure}

Accurate pose estimation is essential for enabling high-precision assembly. This is particularly true for the \texttt{GRASP} primitive in our low-level primitives library, where precise object localization directly affects execution success.
Figure~\ref{fig:pose_estimation} presents some qualitative pose detection results. The high accuracy of the pose estimator enables us to have a high success rate in terms of both the low-level MP policies and the high-level policy.

\subsection{Baseline Description}
To evaluate the effectiveness of our proposed approach, we compare it with several adapted hierarchical baselines. However, due to differences in their design, it is challenging to directly apply their methods. In this section, we provide further details about these baselines, including their structure and the adaptations made to align them with our long-horizon assembly tasks.
\begin{itemize}[leftmargin=1em]
    \item \textbf{MimicPlay}~\cite{wang2023mimicplay}: Since we do not have access to human play data, we retain the hierarchical architecture of MimicPlay but train it using our own expert demonstrations. We modify the high-level policy to predict motion trajectories, and adapt the low-level policy to produce executable actions accordingly. This baseline does not incorporate low-level action primitives.
    \item \textbf{Luo et al.}~\cite{luo2024multi}: We adopt the hierarchical structure from their framework. However, for this baseline, we construct the entire low-level primitive library using imitation learning rather than motion planning or reinforcement learning, which leads to primitives that are less robust and adaptable in diverse scenarios.
\end{itemize}

\begin{table}[]
    \centering
    \caption{Hyperparameters for high-level policy}
    \label{tab:hyper_dit}
    \begin{tabular}{c|c}
    \toprule
        Hyper-parameter & Value \\
        \midrule
        Hidden Dimension & 128 \\
        Number of Blocks & 4 \\
        Number of Heads & 4 \\
        MLP Ratio & 4 \\
        Dropout Prob & 0.1 \\
        \bottomrule
    \end{tabular}
\end{table}
\end{document}